\documentclass[letterpaper, 10 pt, conference]{ieeeconf}  

\IEEEoverridecommandlockouts                              

\title{\LARGE \bf
DriveGen: Towards Infinite Diverse Traffic Scenarios with Large Models
}

\author{Shenyu Zhang$^{1}$, Jiaguo Tian$^{1}$, Zhengbang Zhu$^{1}$, Shan Huang$^{2}$, Jucheng Yang$^{2}$ , Weinan Zhang$^{1}$
\thanks{$^{1}$Dept. of Computer Sci. and Eng., Shanghai Jiao Tong University, Shanghai, 200240, China.}%
\thanks{$^{2}$AI Laboratory, Chongqing Changan Automobile Co. Ltd, Chongqing, 400023, China}%
}

\usepackage{xspace}
\newcommand{\method}{DriveGen\xspace}
\usepackage{graphicx}
\usepackage{subfigure}
\usepackage{amssymb}
\usepackage{amsmath}
\usepackage{caption}
\usepackage{booktabs}
\begin{document}

\maketitle
\thispagestyle{empty}
\pagestyle{empty}

\begin{abstract}

Microscopic traffic simulation has become an important tool for autonomous driving training and testing. Although recent data-driven approaches advance realistic behavior generation, their learning still relies primarily on a single real-world dataset, which limits their diversity and thereby hinders downstream algorithm optimization. In this paper, we propose \method, a novel traffic simulation framework with large models for more diverse traffic generation that supports further customized designs. \method consists of two internal stages: the initialization stage uses a large language model and retrieval technique to generate map and vehicle assets; the rollout stage outputs trajectories with selected waypoint goals from a visual language model and a specifically designed diffusion planner. Through this two-staged process, \method fully utilizes large models' high-level cognition and reasoning of driving behavior, obtaining greater diversity beyond datasets while maintaining high realism. To support effective downstream optimization, we additionally develop \method-CS, an automatic corner case generation pipeline that uses failures of the driving algorithm as additional prompt knowledge for large models without the need for retraining or fine-tuning. Experiments show that our generated scenarios and corner cases have superior performance compared to state-of-the-art baselines. Downstream experiments further verify that the synthesized traffic of \method provides better optimization of the performance of typical driving algorithms, demonstrating the effectiveness of our framework.

\end{abstract}

\section{INTRODUCTION}


Microscopic traffic simulation has become an essential tool for autonomous vehicle (AV) research \cite{c1} as it provides a realistic, low-cost, and efficient training and verification platform for autonomous driving algorithms. Despite advances in physics simulation, mainstream driving simulators \cite{c12,c13,c14} rely primarily on real-world experience replay or heuristics-based controllers to simulate driving behaviors. However, the finite-sized real-world dataset limits the diversity and scalability of available scenarios, and the fidelity of controllers is insufficient.

To overcome the above challenges, learning-based methods are proposed to generate realistic behaviors from real-world driving logs \cite{c15}. However, two major difficulties remain to prevent the wide use of these methods. First, existing methods are mostly learned from a unitary real-world dataset to reproduce high-probability behaviors without further integration of driving knowledge and world understanding, which limits their ability to generalize diverse traffic and customize unseen scenarios. This also leads to a discrepancy between the training performance of autonomous driving algorithms on generated data and their performance in real-world scenarios. Second, efficient construction of corner cases remains a hurdle to previous methods. Corner cases \cite{c16, c17} refer to rare or unusual scenarios that pose significant challenges to the safety and functionality of autonomous vehicles, potentially leading to algorithm failures or collisions. Although several safety-critical scenario construction methods exist \cite{c18}, they are designed to target specific policies. As a result, the model necessitates retraining or fine-tuning when encountering a new policy, thereby limiting its efficiency in practical optimization applications.


Recent advancements in large models (LMs)\footnote[1]{The large models (LMs) mentioned in this paper include both large language models (LLM) and visual language models (VLM).} \cite{c19, c20} offer promising potential to address these difficulties. LMs have demonstrated superior performance in world understanding and can be utilized to deal with complex reasoning tasks with proper prompt instructions. Furthermore, the usage of LMs has been proven effective in end-to-end AV algorithms \cite{c21,c22}, indicating that LMs can provide a comprehensive understanding of driving behavior and traffic simulation.

Inspired by this, we present a novel autonomous driving simulation framework with LMs called \method. Leveraging generative capabilities of driving behavior and traffic provided by state-of-the-art LMs, \method supports synthesized data generation
with greater diversity beyond a single dataset while maintaining high realism. In specific, it divides microscopic traffic simulation into two stages: \textit{scenario initialization} and \textit{scenario rollout}. The initialization stage uses a large language model and retrieval technique to generate map and vehicle assets; while the rollout stage outputs dynamic trajectories through selected waypoint goals from a visual language model and a specifically designed diffusion planner. Relying on this pipeline, \method is capable of generating more useful traffic scenarios for downstream algorithm optimization.



To support efficient corner case generation, we further propose \method-CS (\textbf{C}orner ca\textbf{S}e). \method-CS takes features from collision cases of the targeted driving algorithm as additional knowledge prompts to help LMs generate corner cases where the algorithm tends to fail. This prompt-based architecture allows the method to be generalized on different algorithms without additional training or finetuning costs, making it more suitable for application needs.
\begin{figure*}[ht]
    \centering 
    \includegraphics[width=0.8\linewidth]{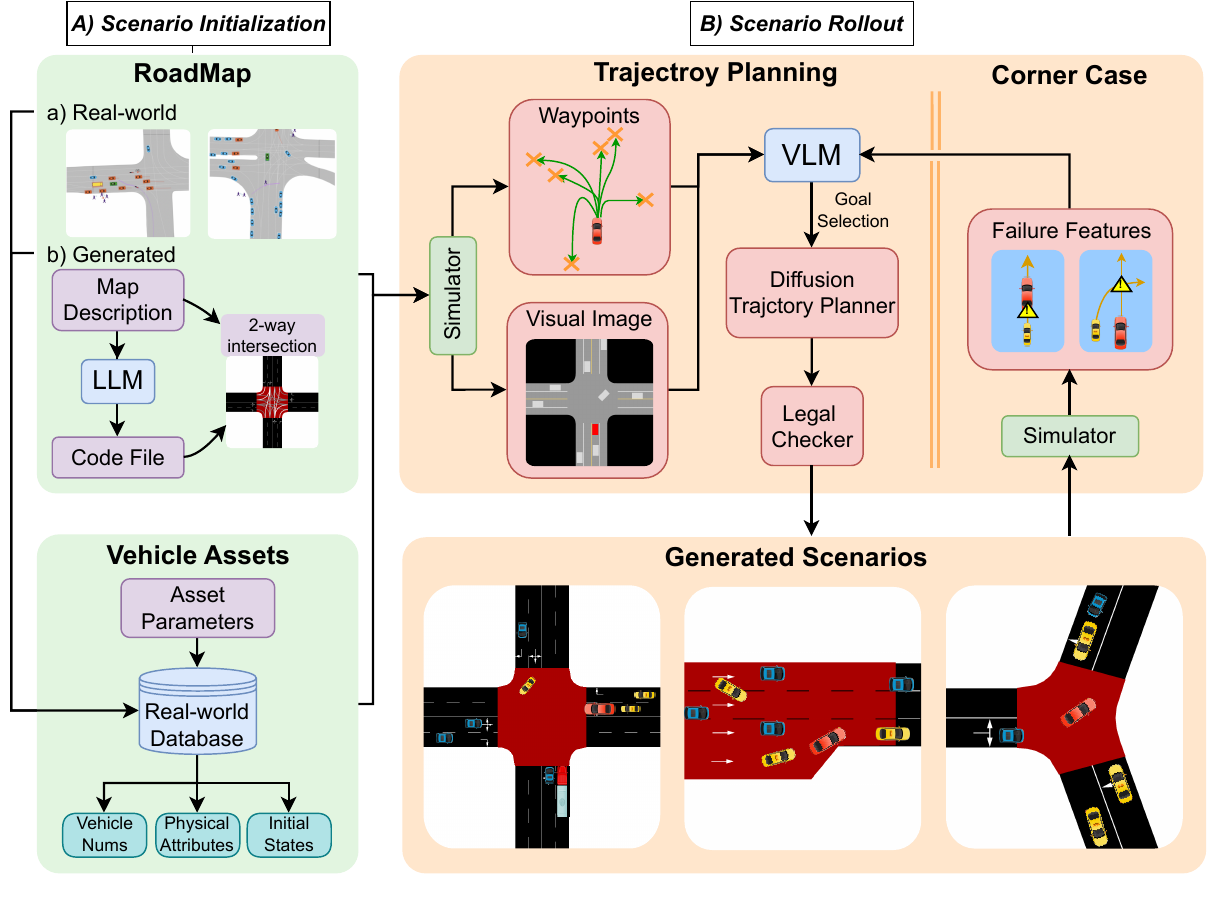}
    \caption{DriveGen Pipeline. \method consists of two internal stages: A) scenario initialization and B) scenario rollout.  \method-CS, shown in the rollout stage, is an extension module of DriveGen, which utilizes failure scenarios of the driving algorithm to form knowledge prompts that further guide the VLM in generating corner cases.} 
    \label{fig-method}
\end{figure*}

In summary, the contribution of this paper is threefold:
\begin{itemize}
    
    \item We introduce \method, a novel two-stage traffic simulation framework with large models that generates traffic scenarios with greater diversity while maintaining high realism.
    \item We propose \method-CS, an efficient corner case generation approach. \method-CS enables automatic generation by understanding failures and summarizing deficiencies of driving algorithms and can be deployed on different algorithms without additional training or finetuning costs.
    \item The experiments demonstrate the superior performance of \method and \method-CS in scenario quality, performance of downstream tasks, and evaluation of corner cases.
\end{itemize}

\section{Related Works}
\textbf{Microscopic Traffic Simulation.} 
Recent works have explored data-driven techniques, modeling driving behavior with deep neural networks and generative models. For example, BITS \cite{c10} proposes to generate traffic by exploiting the bi-level hierarchy of high-level intent inference and low-level driving behavior imitation. CTG \cite{c35} leverages diffusion modeling and differentiable logic to guide generated trajectories to meet traffic rules while maintaining realism. TrafficGen \cite{c32} is an autoregressive neural generative model with an encoder-decoder architecture. RealGen \cite{c54} proposes a retrieval augmented generation framework with a contrastive autoencoder model for controllable driving scenario generation. TrafficRLHF \cite{c57} improves the realism of current methods at the cost of additional preference dataset construction without consideration of diversity. These methods enhance overall performance through various techniques that depict the characteristics of driving behavior, but the heavy reliance on training datasets limits their abilities to generalize diverse and unseen scenarios or support customized scenario definitions.

\begin{figure*}[ht]
    \centering 
    \includegraphics[width=1.0\linewidth]{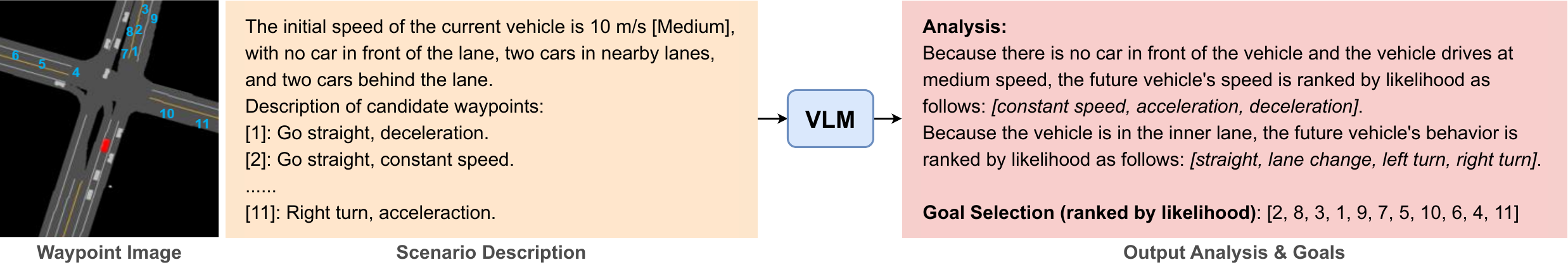}
    \caption{Example Prompt for Goal Selection. We show partial text for brevity.} 
    \label{fig-prompt}
\end{figure*}

\textbf{Large Models (LMs) in Autonomous Vehicles.} Since LMs demonstrate their superior performance in complex reasoning and world understanding, utilizing them to assist in autonomous vehicles has now become a new trend. LMs have proven efficiency in autonomous driving strategies by providing high-level motion commands \cite{c21,c22,c50,c51,c52}. Several papers have investigated the potential of using LMs in traffic simulation. For example, ChatScene \cite{c56} generates textually described scenarios using large language models to facilitate safety-critical scenarios with control commands from CARLA. Most similar work to ours is LCTGen \cite{c53}, which utilizes a large language model (LLM) to learn a heuristic intermediate representation of scenarios from language
inputs for motion generator. Compared to LCTGen, our \method enables map generation and uses visual information to drill the visual language model (VLM) into the core of trajectory planning beyond representation, making further enhancement toward diverse and realistic traffic simulation.

\textbf{Corner Case Generation.} There are three main categories of algorithms for corner case generation. First, data-driven approaches \cite{c36,c37} utilize real-world experiences to generate testing scenarios. However, it is challenging due to the unbalanced nature of collected data, as risky scenarios are very limited. Given that, it is more efficient to add safety-critical objects such as adversarial vehicles to attack the driving algorithm, which forms the adversarial method \cite{c38,c39}. Knowledge-driven corner case generation \cite{c40} investigates how to integrate human prior knowledge and traffic rules to guide the generation, but it is difficult to represent such rules formally. We observe that the majority of the aforementioned algorithms share a common limitation: they require retraining or fine-tuning on driving data from the updated algorithm, which significantly reduces their efficiency during the iterative optimization process of the targeted algorithm. Our \method-CS utilizes statistical features from collision scenarios of the target algorithm as knowledge prompts to generate corner cases, thereby eliminating the need for additional retraining or fine-tuning.




\section{Methodology}
\method is a microscopic traffic simulation framework that utilizes the generative capabilities of the latest foundation models to generate diverse and realistic traffic scenarios. We illustrate the whole pipeline in Figure \ref{fig-method}, composed of two internal stages: \textit{scenario initialization} and \textit{scenario rollout}, while \method-CS is further explained to the right of the double vertical lines in the rollout stage.

\subsection{Problem Formulation}

A complete microscopic traffic scenario consists of both static and dynamic elements. Static elements includes semantic map $M_s$ and attributes $P = \{p_0, \cdots, p_N\}$ of 
$N$ traffic participants\footnote[2]{In this paper, we focus our discussion on vehicles.}. The attributes of a vehicle contain its type, 3d size, and initial state (position, velocity, heading) at the first frame in the simulation. The dynamic elements include traffic signals $M_d$ in the map and vehicle trajectories $T = \{\tau^0, \tau^1, \cdots, \tau^N\}$. We define the total duration of the traffic as $T_s$. The trajectory of $i$-th vehicle is formed by a sequence of its states in all timesteps $\tau^i = \{s^i_0, s^i_1, \cdots, s^i_{T_s}\}$. We use a scenario-level coordinate to achieve a consistent representation of trajectories. Given that the traffic signals $M_d$ are controlled by simulators, the key components of traffic scenario $S$ can be described as $S = \{M_s, P, T\}$. The autonomous driving algorithm can then be trained and evaluated by replacing the target vehicle in generated scenarios.




\subsection{Scenario Initialization}

In the initialization stage, we describe the generation process of semantic map $M_s$ and vehicle assets $P$.

\textbf{Roadmap Generation.} 
Most previous data-driven approaches model only driving behaviors and neglect to study the semantic map. When researchers want to optimize policy performance in specific road structures, filtering out such maps from the dataset is often time-consuming (no label for maps) and not necessarily sufficient in number. \method enables map customization through a well-designed knowledge prompt even for non-experts, and the generated maps can be directly loaded into popular simulators.

To achieve this, we select the popular SUMO network definitions in XML format to construct the map. The map code consists of three parts: \textit{Lane} denotes the vehicle lane; \textit{Junction} represents the connection between the lanes; and \textit{Connection} defines the legal route for vehicles. We equip the LLM with rule-based knowledge prior and typical examples (e.g. intersection, roundabout, highway), making it generate the map file based on text description with a high success rate. The map can also be refined with human feedback for LLM. For example, modify the location where the ramp merges into the main road. The framework accepts real-world maps as well.

\textbf{Vehicle Assets Generation.}
The static map then serves as a basis upon which the method can condition and subsequently place dynamic vehicles in the traffic scenario. The generation process defines the following things in order: the number of vehicles, the physical attributes of the vehicles, and the initial state $s_0$ of the vehicles. Here we collect vehicle information from the real-world dataset to build a database. Once the number and type of vehicle are determined, we retrieve the corresponding vehicles from the database. The first one is used as the target vehicle and the rest is used as social vehicles. To determine the initial states, we first randomly place the target vehicle in a feasible area according to the map and give it the correct heading along the lane and initial speed. Then social vehicles are placed to surround the target vehicle with similar speed to create rich interaction.

\subsection{Scenario Rollout}
Given the map and the initial states of vehicles, we generate diverse and realistic trajectories by leveraging image information simulated by the simulator and VLM, thereby constructing high-quality scenarios. Direct querying for low-level continuous trajectories leads to poor performance as it fails to fully leverage the high-level cognitive and reasoning capabilities of LMs. Instead, we let VLM make discrete decisions based on well-structured information. In detail, we conduct a rule-based path retrieval method that processes waypoints produced by the simulator. It first searches for all feasible routes and then samples reasonable goals considering the initial speed and heading of a vehicle. Such candidates are marked by numbers in a 200m$\times$200m vehicle-centric bird-eye image that shows surrounding road structure and vehicles. We then send the image to VLM together with the scenario description. The VLM outputs reasonable goals $s_{T_s}$ ranked by likelihood for each vehicle given prompt guidance. See Figure \ref{fig-prompt} for a detailed example. Leveraging a high-level understanding of driving behavior from the large model, we found that this approach is far superior to random selection in realism and ensures greater diversity through various waypoints under different routes.


\begin{figure}[h]
    \centering 
    \includegraphics[width=0.9\linewidth]{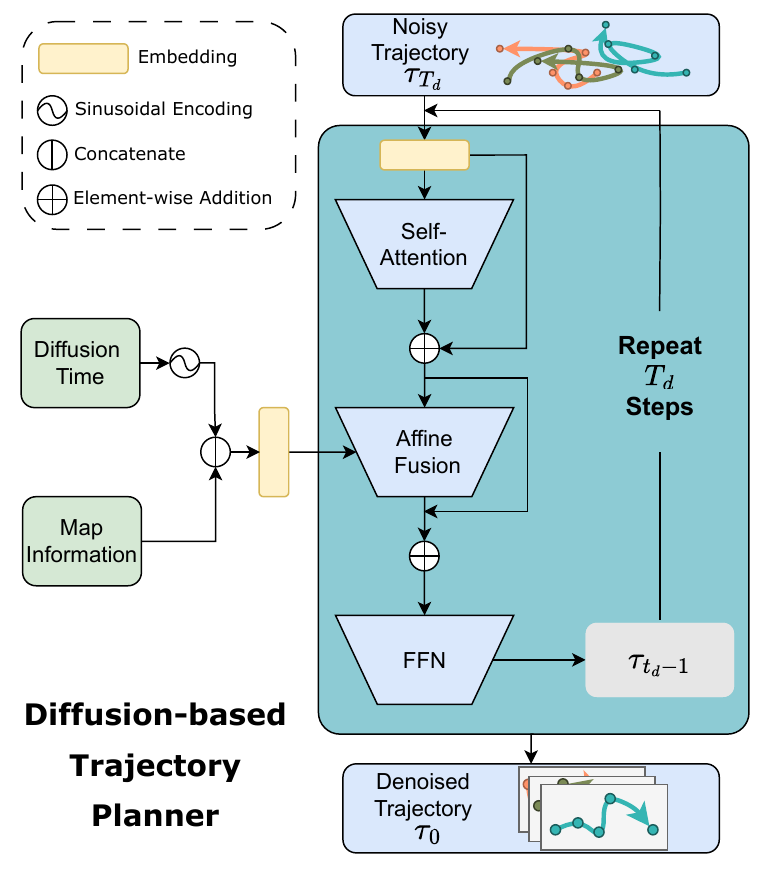}
    \caption{DriveGen Data-driven Planner. The planner combines the transformer network with diffusion models to capture temporal and multi-modal driving behavior features, thus recovering realistic and diverse trajectories. } 
    \label{fig-planner}
\end{figure}

To fill realistic trajectories like human drivers, we implement a novel data-driven diffusion-based trajectory planner to capture human driving style. The diffusion model learns $T_d$ steps reverse denoising process that recovers the original trajectory distribution $\tau_0$ from pure Gaussian noise $\tau_{T_d}$. With the initial state and final goal of the vehicle, the optimization object can be written as: 
\begin{equation}
    \mathcal{L} = 
  \mathbb{E}_{\tau_0,t_d}[\| \epsilon -\epsilon_\theta(\tau_{t_d},s_0,s_{T_s},t_d) \|^2],
\end{equation}
where $\epsilon_\theta$ is the parameterized model that learns to fit noise $\epsilon$ in each denoising step $t_d$.  
 The detailed network is illustrated in Figure \ref{fig-planner}. We add a fusion module between the self-attention block and feed-forward network (FFN) of the transformer, which incorporates diffusion noise embedding $X_{t_d}$ with map information $m$ and diffusion timestep $t_d$ by deep affine transformation: 
\begin{equation}
    \mathrm{B} = \phi_B(m,t_d), \mathrm{W} = \phi_W(m,t_d), X_{t_d}' = X_{t_d} \cdot \mathrm{W} + \mathrm{B},
\end{equation}
where $\phi_B,\phi_W$ is a multi-layer perception (MLP) and $(\cdot)$ denotes Hadamard product. This combination enables the model to capture temporal and multi-modal driving behavior features, thus recovering realistic trajectories. 

We generate the trajectory in the same order as vehicle assets. Each trajectory is examined through a legal checker to judge whether it conforms to traffic rules, avoids collisions with other vehicles, and is feasible within dynamic constraints.
The unqualified trajectory will be regenerated by the planner. Finally, the collection of all trajectories and the map form a complete traffic scenario.


\subsection{\method-CS}
To support efficient algorithm optimization, we further propose \method-CS, a collaboration between \method and \textbf{C}orner ca\textbf{S}e generation techniques. This paradigm utilizes collision scenarios of the targeted algorithm as extra domain knowledge for LMs to select more threatening social vehicle trajectories for the targeted algorithm.

\textbf{Processing Failed Examples.} As mentioned above, corner cases pose significant challenges to the safety of AV. So the collision scenarios of the driving algorithm are the best corner cases. The collision scenarios are first collected through evaluation on generated \method scenarios where the target vehicle controlled by the algorithm collides with a social vehicle. Then we extract their relative speed, heading, and driving behavior when collision happens as features. Through statistics, we can find feature values with the highest collision probability, and send the corresponding information and visual trajectory collision images to the VLM as additional knowledge prompts for generation.



\begin{table*}[ht]
\caption{Scenario quality evaluation. We compute the ratio of methods to the real dataset for the diversity metric, with the values in parentheses representing the relative error. `$\rightarrow$' means results are better if the metric is closer to the real dataset.}
\label{table_realism}
\begin{center}
\begin{tabular}{c c c c c c}
\toprule
& min-sADE$\downarrow$ & min-sFDE$\downarrow$ & mean-sADE$\downarrow$ & mean-sFDE$\downarrow$ & diversity$\rightarrow$\\
\midrule
SceneTransformer \cite{c42}  & 1.50 & 3.12 & 2.92 & 3.89 & 0.61 (-0.39)\\ 
BITS \cite{c10} & 1.24 & 2.87 & \textbf{1.92} & \textbf{3.42} & 0.57 (-0.43) \\ 
TrafficGen \cite{c32} & 3.23 & 5.26 & 5.21 & 8.89 & 0.75 (-0.25) \\ 
LCTGen \cite{c53} & 1.89 & 2.64 & 2.78 & 4.10 & 0.76 (-0.24) \\ 
\midrule
DriveGen w/o gs  & 2.34 & 3.62 & 4.15 & 6.21 & 1.20 (+0.20) \\ 
DriveGen  & \textbf{1.18} & \textbf{2.16} & 2.59 & 3.64 & \textbf{0.93 (-0.07)} \\ 
\bottomrule
\end{tabular}
\end{center}
\end{table*}
\textbf{Corner Case Generation.} 
Differing from \method, the initial position of the target vehicle (controlled by algorithm) is additionally identified in the given RGB image for social vehicles, and we require the VLM to select goals that have a higher probability of making the targeted algorithm fail according to the given knowledge. Since the candidate waypoints follow traffic rules and the trajectory planner learns from human distribution, the generated corner cases maintain realism and plausibility. 


\subsection{Implement Details}
We use Claude-3-opus \cite{c41} as LLM to generate the map code file and GPT-4-turbo \cite{c55} as VLM to select proper goals for vehicles. For the diffusion trajectory planner, we encode 256$\times$256 vehicle-centric rasterized map by a 4-layer CNN as map information. For the main backbone, We clone a 3-layer transformer with an embedding dimension of 256 and 4 heads, then insert an affine fusion block with two 2-layer MLPs as seen in Figure~\ref{fig-planner}. We train the model for 90 epochs with a learning rate of 1e-4, batch size of 32, and diffusion model steps of 100.

\section{Experiment}

We seek to validate those primary claims in experiments:
\begin{itemize}

\item \method achieves superior diversity in microscopic traffic simulation while maintaining high realism.

\item \method provides better performance improvement compared to the previous method to support the training of the downstream driving algorithm.



\item \method-CS generates qualified corner cases that detect the weakness of the targeted algorithm without additional training or finetuning cost.
\end{itemize}

\begin{table*}[h]
\caption{Downstream optimization performance. }
\label{table_downstream}
\begin{center}
\begin{tabular}{c c c c c c c c c}
\toprule
& \multicolumn{4}{c|}{GAIL} & \multicolumn{4}{c}{SAC} \\
& CR$\downarrow$ & Dis$\uparrow$ & Comfort$\uparrow$ & \multicolumn{1}{c|}{Div$\uparrow$} & CR$\downarrow$ & Dis$\uparrow$ & Comfort$\uparrow$ & Div$\uparrow$ \\
\midrule
Dataset & 0.15 & 55.67 & 4.59 & 8.67 & 0.18 & 48.27 & 1.49 & 5.32 \\
\midrule
SceneTransformer \cite{c42} & 0.16 & 52.45 & 5.26 & 7.49 & 0.19 & 42.72 & 1.21 & 3.85 \\
BITS \cite{c10}& 0.15 & 56.90 & 5.10 & 7.84  & 0.21 & 49.36 & 1.75 & 3.07\\
TrafficGen \cite{c32} & 0.14 & 58.30 & 4.95 & 8.29 & 0.20 & 52.25 & 2.10 & 5.35 \\
LCTGen \cite{c53} & 0.13 & 60.31 & 5.31 & 8.75 & 0.18 & 51.77 & 2.58 & 6.01\\
\midrule
\method w/o m & 0.13 & 61.32 & 5.56 & 9.10 & 0.17 & 51.89 & 2.24 & 6.53\\
\method w/o gs & 0.15 & 55.90 & 5.21 & 8.65 & 0.20 & 51.37 & 2.34 & 5.78\\
\method & \textbf{0.12} & \textbf{64.73} & \textbf{6.25}  & \textbf{10.09} & \textbf{0.15} & \textbf{55.89} & \textbf{3.70} & \textbf{7.77}\\
\bottomrule
\end{tabular}
\end{center}
\end{table*}

\subsection{Experiment Setup}
\textbf{Dataset.} We use the large-scale real-world Argoverse2 Motion Forecasting Dataset. Each scenario is simulated at 10Hz with $T_s$ = 100 timesteps, making a 10-second multi-agent traffic. We limit the maximum number of vehicles to $N$ = 16, which consists of a targeted vehicle and 15 nearest social vehicles at the initial frame. We train our diffusion planner and other baselines on its 187k train split and test on its 25k validation split.

\textbf{Simulation Environments.}
In our experiments, we use SMARTS \cite{c47} as the traffic simulator.
Note that \method is independent of the used simulator and can be integrated into any other simulator with multi-agent traffic flow simulation and bird-eye view rendering capabilities.

\subsection{Scenario Quality Evaluation}
We evaluate the quality of \method's generated scenarios by comparing them to real-world scenarios from the dataset. To make a fair comparison, we generate trajectories $T$ conditioned on the same map and initial states$(M_s, P)$ of real-world scenarios for all methods.  

\textbf{Metrics.} To measure the realism of generated trajectories, we employ the standard average distance error (ADE) and final distance error (FDE) between them and real-world trajectories. For each test scenario, we sample $s$ = 10 generated scenarios and calculated the minimum and average distance error across samples to get four fine-grained statistics: min-sADE, min-sFDE, mean-sADE, and mean-sFDE. We describe the diversity of the generated scenarios following \cite{c48}. We pick the sample with the best min-sADE as representative of each test scenario to form the generated dataset. For each scenario in the dataset, we then transform the trajectory of each vehicle based on its initial position and heading to the origin of its coordinate frame. After that, the dataset is randomly split into pairs and the average joint state difference $D(S_1, S_2)$ in each pair is calculated. Here $s$ refers to state and $S$ refers to scenario. 
\begin{equation}
\begin{aligned}
    & d(\tau_1, \tau_2) = \Sigma_{i=1}^{T_s} ||s_i^1 - s_i^2||^2, \\
    & D(S_1, S_2) = \Sigma_{i=1}^N\Sigma_{j=1}^N d(\tau_i^1, \tau_j^2).
\end{aligned}
\end{equation}
The average difference of all pairs serves as the diversity of the dataset. We take the diversity of the real-world dataset as the baseline, denoted as 1, and compute the ratio between the generated dataset and the real dataset to quantify the error magnitude.

\textbf{Baselines.} We consider recent state-of-the-art data-driven diverse traffic generation methods as baselines. SceneTransformer \cite{c42} is a unified multi-agent trajectory prediction architecture for multi-modal output. BITS \cite{c10} generates traffic behaviors by exploiting the bi-level hierarchy of high-level intent inference and low-level driving behavior
imitation. TrafficGen \cite{c32} is an autoregressive neural generative model with an encoder-decoder architecture. LCTGen \cite{c53} learns a heuristic intermediate representation of scenarios from language inputs for trajectory generator. We skip the vehicle placement module of TrafficGen and the map retrieval module of LCTGen since maps and initial states are given. We divide SceneTransformer and BITS into prediction baselines as they focus on learning driving behavior while considering TrafficGen and LCTGen as generative baselines as they focus on scenario generation.

\begin{table*}[ht]
\caption{Corner case performance. }
\label{table_cornercase}
\begin{center}
\begin{tabular}{c c c c c c c c c c c c c}
\toprule
& \multicolumn{4}{c|}{IDM} & \multicolumn{4}{c|}{GAIL} & \multicolumn{4}{c}{SAC} \\
& CR$\uparrow$ & Dis$\downarrow$ & Comfort$\downarrow$ & \multicolumn{1}{c|}{Div$\downarrow$} & CR$\uparrow$ & Dis$\downarrow$ & Comfort$\downarrow$ & \multicolumn{1}{c|}{Div$\downarrow$} & CR$\uparrow$ & Dis$\downarrow$ & Comfort$\downarrow$ & \multicolumn{1}{c}{Div$\downarrow$} \\
\midrule
Dataset & 0.07 & 49.04 & 0.325 &  15.76 & 0.15 & 55.67 & 4.59 & 8.67 & 0.18 & 48.27 & 1.49 & 5.32\\
\midrule
AdvSim \cite{c44}& 0.11 & 41.60 & 0.330 & 14.25 & 0.23 & 35.41 & 3.85 & \textbf{5.01} & 0.37 & 26.64 & 1.72 & 5.14 \\ 
DiffScene \cite{c45} & 0.14 & 44.86 & 0.303 & 15.75 & 0.26 & 38.27 & 4.72 &  5.53 & \textbf{0.41} & 27.56 & 1.51 & 4.39 \\
\midrule
\method-CS & \textbf{0.25} & \textbf{39.50} & \textbf{0.246} & \textbf{4.43} & \textbf{0.28} & \textbf{34.72} & \textbf{2.51} & 6.40 & 0.39 & \textbf{25.53} & \textbf{1.13} & \textbf{3.62}\\ 
\bottomrule
\end{tabular}
\end{center}
\end{table*}
\textbf{Results.} The quantitative results from Table~\ref{table_realism} demonstrates superior performance of \method on scenario quality. For realism, \method achieves the best min-sADE and min-sFDE, which significantly outperforms generative baselines. Importantly, \method makes substantial improvement on diversity when generating large-scale scenario dataset, which obtains 50\%+ better performance over prediction baselines and 20\%+ over generative baselines in absolute terms and 
achieves 3x smaller relative error than all baselines. We note that \method obtains greater performance in mean-sADE and mean-sFDE over generative methods but worse than BITS. We conjecture that this is because generation progress with diverse goals selected by VLM leads to a slight loss of consistency. However, this consistency is demonstrated unhelpful when applied for downstream optimization below.

\subsection{Downstream Optimization}
Eventually, the generated scenarios will be used by the autonomous driving algorithm as training or evaluation samples. So it is important to measure the effectiveness of generated scenarios by conducting an optimization experiment. We choose GAIL and SAC as our driving algorithms. GAIL is a typical imitation learning algorithm that learns driving behavior from expert data
while SAC is a popular reinforcement learning algorithm that maximizes accumulated rewards from traffic scenarios. We parameterized the driving policy with the same 2-layer MLP with a latent dimension of 256. The policy is first trained on the Argoverse2 train split and finetuned on generated scenarios to compare the improvement. The algorithm's performance is evaluated by replacing the target vehicle with the learned policy in test scenarios.

\textbf{Baselines.} We use the same generated scenario dataset as when calculating diversity above. Additionally, for generative baselines, we replace 25\% of the data with scenarios generated using the complete method (new vehicle assets for TrafficGen, LCTGen; new maps and vehicle assets for \method).

\textbf{Metrics.} We consider multifaceted performance evaluation metrics from AV. We measure the driving capability through the average collision rate (CR) and driving distance (Dis) of scenarios. We take the reciprocal of average linear acceleration used in \cite{c49} to represent the driving comfort (Comfort) and the average standard deviation of trajectories in test scenarios to measure the diversity (Div) of learned driving styles. The better the algorithm performs on these metrics, the more effective the scenarios provided.

\textbf{Results.} We present all evaluations for downstream tasks in Table~\ref{table_downstream}. The results show that scenarios of \method provide the most effective enhancement to driving policy performance in downstream optimization, offering the greatest benefit for autonomous vehicle algorithm learning. For both GAIL and SAC, finetuning on DriveGen scenarios results in superior performance in all metrics, suggesting that our generated scenarios can help the algorithm learn safe, comfortable, and diverse driving skills. Different from the consistent improvement brought by the generative baselines in metrics, the prediction baselines even bring negative effects, especially in diversity where both algorithms suffer performance decreases. This further verifies the importance of diversity in traffic simulation and the improvement significance brought by our method.

\subsection{Corner Case Quality}
We evaluate the quality of DriveGen-CS's corner cases comprehensively by doing tests on them with different driving algorithms. Here we additionally add IDM, a heuristic-based controller for driving, acting according to a scheduled car-following logic. The policies of GAIL and SAC are directly obtained from the downstream experiments trained on the real-world dataset. These three algorithms represent three different driving styles. IDM stands for regularized control, GAIL stands for human-like control, and SAC stands for reward-driven control, which has a more aggressive driving strategy as it is rewarded by driving distance. All the algorithms are tested on 1k generated corner cases.

\textbf{Baselines.} We choose two typical corner case generation methods. AdvSim \cite{c44} is an adversarial framework to generate safety-critical scenarios that modiﬁes the actors’ trajectories in a physically plausible manner. DiffScene \cite{c45} designs several adversarial optimization objectives to guide the diffusion generation under predefined adversarial budgets to ensure the generation of safety-critical scenarios. We first train these two methods on real-world dataset and then finetuned them with traffic scenarios controlled by the targeted algorithm.

\textbf{Metrics.} We use the same performance evaluation metrics as for the downstream task. Differing from the downstream experiment, the poorer performance of driving policies indicates the higher quality of corner cases as their purpose is to challenge the performance of the targeted algorithm. We also recorded the GPU hours required to generate 1k scenarios for each method.

\begin{table}[ht]
\begin{center}
\caption{GPU hours for corner case generation. }
\label{table_gpucost}
\begin{tabular}{c c c}
    \toprule
     & Finetune & Generation \\
     \midrule
    AdvSim \cite{c44} & 4.2 & \textbf{0.35} \\
    DiffScene \cite{c45} & 5.7 & 0.89 \\
    \midrule
    \method-CS & \textbf{0} & 0.75 \\
    \bottomrule
\end{tabular}
\end{center}
\end{table}

\textbf{Results.} The corner case results are shown in Table~\ref{table_cornercase}. In most of the metrics of the three algorithms, \method-CS has achieved leading performance, indicating that its corner cases are more challenging and qualified. For IDM, \method-CS obtains a much more notable increase in collision rate than other baselines, and most of them produce incoming traffic in the adjacent lane, demonstrating a clear detection of IDM's weakness. This indicates that \method-CS can make better attacks at algorithms that obey certain rules. Another significant advantage of \method-CS is shown in Table~\ref{table_gpucost}. By using VLM to analyze failure scenarios to find algorithm weaknesses, \method-CS avoids the GPU training cost caused by the parameter change of algorithms and leaves with only inference cost in the generation procedure, making an 80\% reduction in GPU hours. This property makes \method-CS more suitable for iterative training of autonomous driving algorithms.

\section{Ablation Study}
To verify the effectiveness of using LMs, we mask out the map generation module and the goal selection module respectively, referred to as \method w/o m and \method w/o gs.

\textbf{Customized Map Generation.} Without new maps generated from descriptive text, only vehicle assets are further generated to make diverse scenario initialization. As a result, performance has declined on all metrics, as seen in Table~\ref{table_downstream}. This indicates that the introduction of new customized maps can enrich the scenario diversity of the dataset and effectively improve the optimization performance of downstream algorithms.

\textbf{Trajectory Planning.} In \method w/o gs, we replace the selection of goals from VLM to randomization. From the results in Table~\ref{table_realism}, the random goals lead to significant performance loss in all four realistic metrics. Although the diversity has increased in absolute terms, it has greatly deviated from the real-world dataset. Furthermore, the performance loss in downstream optimization in Table~\ref{table_downstream} shows that high diversity is not helpful for training without a guarantee of realism. This suggests that VLM plays a crucial role in making diverse trajectories while maintaining realism.


\section{Conclusion}
We present \method, a novel traffic simulation framework with large models. \method leverages LMs' high-level cognition and reasoning of driving behaviors and traffic, enabling it to support customized design in maps, vehicle assets, and trajectories, and thus generates diverse and realistic traffic scenarios. The experiments demonstrate our generated scenarios notably surpass state-of-the-art baselines in diversity and provide improved optimization performance for autonomous driving algorithms. We further develop \method-CS, a corner case generation extension that can be generalized on different algorithms without additional training or finetuning costs.


\begin{thebibliography}{99}
\bibitem{c1} N. Kalra and S. M. Paddock, Driving to Safety: How Many Miles of Driving Would It Take to Demonstrate Autonomous Vehicle Reliability? RAND Corporation, 2016. [Online]. Available: http://www.jstor.org/stable/10.7249/j.ctt1btc0xw

\bibitem{c2} Waymo, “Safety report,” Waymo LLC, 2021, Available at https://waymo.com/safety/safety-report. Retrieved on July 4, 2021.

\bibitem{c3} Uber Advanced Technologies Group, “A principled approach to safety,” 2020, Available at https://uber.app.box.com/v/UberATGSafetyReport.

\bibitem{c4} NVIDIA, “Self-driving safety report,” 2021, Available at https://images.nvidia.com/content/self-driving-cars/safety-report/auto-print-self-driving-safety-report-2021-update.pdf.

\bibitem{c5} Argo AI, “Developing a self-driving system you can trust,” Apr.2021, Available at https://www.argo.ai/wp-content/uploads/2021/04/ArgoSafetyReport.pdf.

\bibitem{c6} P. A. Lopez, M. Behrisch, L. Bieker-Walz, J. Erdmann, Y.-P. Fl¨otter¨od, R. Hilbrich, L. L¨ucken, J. Rummel, P. Wagner, and E. Wießner, “Microscopic traffic simulation using sumo,” in International Conference on Intelligent Transportation Systems (ITSC), 2018, pp. 2575–2582.

\bibitem{c7} J. Casas, J. L. Ferrer, D. Garcia, J. Perarnau, and A. Torday, “Traffic simulation with aimsun,” in Fundamentals of traffic simulation. Springer, 2010, pp. 173–232.

\bibitem{c8} M. Fellendorf and P. Vortisch, “Microscopic traffic flow simulator vissim,” in Fundamentals of traffic simulation. Springer,2010, pp.63–93.

\bibitem{c9} S. Suo, S. Regalado, S. Casas, and R. Urtasun, “Trafficsim: Learning to simulate realistic multi-agent behaviors,” in Proceedings of the IEEE/CVF Conference on Computer Vision and Pattern Recognition(CVPR), June 2021, pp. 10 400–10 409.

\bibitem{c10} D. Xu, Y. Chen, B. Ivanovic, and M. Pavone, “Bits: Bi-level imitation for traffic simulation,” 2022.

\bibitem{c11} A. Dosovitskiy, G. Ros, F. Codevilla, A. Lopez, and V. Koltun, “CARLA: An open urban driving simulator,” ser. Proceedings of Machine Learning Research, S. Levine, V. Vanhoucke, and K. Goldberg, Eds., vol. 78, 13–15 Nov 2017, pp. 1–16. [Online]. Available: http://proceedings.mlr.press/v78/dosovitskiy17a.html

\bibitem{c12} Dosovitskiy A, Ros G, Codevilla F, et al. CARLA: An open urban driving simulator[C]//Conference on robot learning. PMLR, 2017: 1-16.

\bibitem{c13} Li Q, Peng Z, Feng L, et al. Metadrive: Composing diverse driving scenarios for generalizable reinforcement learning[J]. IEEE transactions on pattern analysis and machine intelligence, 2022, 45(3): 3461-3475.

\bibitem{c14} Gulino C, Fu J, Luo W, et al. Waymax: An accelerated, data-driven simulator for large-scale autonomous driving research[J]. Advances in Neural Information Processing Systems, 2024, 36.

\bibitem{c15} Chen D, Zhu M, Yang H, et al. Data-driven Traffic Simulation: A Comprehensive Review[J]. IEEE Transactions on Intelligent Vehicles, 2024.

\bibitem{c16} Webb, N., Smith, D., Ludwick, C., Victor, T., Hommes, Q.,
Favaro, F., Ivanov, G., and Daniel, T. Waymo’s safety
methodologies and safety readiness determinations. arXiv
preprint arXiv:2011.00054, 2020.

\bibitem{c17} California Department of Motor Vehicle Disengagement
 Report.

\bibitem{c18} Ding W, Xu C, Arief M, et al. A survey on safety-critical driving scenario generation—A methodological perspective[J]. IEEE Transactions on Intelligent Transportation Systems, 2023, 24(7): 6971-6988.

\bibitem{c19} Achiam J, Adler S, Agarwal S, et al. Gpt-4 technical report[J]. arXiv preprint arXiv:2303.08774, 2023.

\bibitem{c20} Zhang J, Huang J, Jin S, et al. Vision-language models for vision tasks: A survey[J]. IEEE Transactions on Pattern Analysis and Machine Intelligence, 2024.

\bibitem{c21} Cui Y, Huang S, Zhong J, et al. Drivellm: Charting the path toward full autonomous driving with large language models[J]. IEEE Transactions on Intelligent Vehicles, 2023.

\bibitem{c22} Tian X, Gu J, Li B, et al. Drivevlm: The convergence of autonomous driving and large vision-language models[J]. arXiv preprint arXiv:2402.12289, 2024.

\bibitem{c23} Krajzewicz D. Traffic simulation with SUMO–simulation of urban mobility[J]. Fundamentals of traffic simulation, 2010: 269-293.

\bibitem{c24} Fellendorf M, Vortisch P. Microscopic traffic flow simulator VISSIM[J]. Fundamentals of traffic simulation, 2010: 63-93.

\bibitem{c25} Casas J, Ferrer J L, Garcia D, et al. Traffic simulation with aimsun[J]. Fundamentals of traffic simulation, 2010: 173-232.

\bibitem{c26} Meiring G A M, Myburgh H C. A review of intelligent driving style analysis systems and related artificial intelligence algorithms[J]. Sensors, 2015, 15(12): 30653-30682.

\bibitem{c27} Lazar H, Rhoulami K, Rahmani D. A review analysis of optimal velocity models[J]. Periodica Polytechnica Transportation Engineering, 2016, 44(2): 123-131.

\bibitem{c28} Z. Zhu et al., “RITA: Boost driving simulators with realistic interactive trafﬁc ﬂow,” in Proc. Int. Conf. Distrib. Artif. Intell., Singapore, 2023, pp. 1–10.

\bibitem{c29} R. P.Bhattacharyya,D.J. Phillips,B.Wulfe,J.Morton,A.Kueﬂer, and M.J. Kochenderfer, “Multi-agent imitation learning for driving simulation,” in Proc. RSJ Int. Conf. Intell. Robots Syst., 2018, pp. 1534 1539.

\bibitem{c30} Y. Cao, B. Ivanovic, C. Xiao, and M. Pavone, “Reinforcement learning with Human feedback for realistic trafﬁc simulation,” 2023, arXiv:2309.00709.

\bibitem{c31} H. Niu et al., “(Re)$^2$H2O: Autonomous driving scenario generationvia reversely regularized hybrid ofﬂine-and-online reinforcement learning,” in Proc. IEEE Intell. Veh. Symp., Anchorage, AK, US, 2023, pp. 1–8.

\bibitem{c32} L. Feng, Q. Li, Z. Peng, S. Tan, and B. Zhou, “Trafﬁcgen: Learning to generate diverse and realistic trafﬁc scenarios,” in Proc. IEEE Int. Conf. Robot. Automat., 2023, pp. 3567–3575.

\bibitem{c33} Z. Zhang, A. Liniger, D. Dai, F. Yu, and L. Van Gool, “TrafﬁcBots: Towards world models for autonomous driving simulation and motion prediction,” 2023, arXiv:2303.04116.

\bibitem{c34} S. Suo et al., “MixSim: A hierarchical framework for mixed reality trafﬁc simulation,” in Proc. IEEE/CVF Conf. Comput. Vis. Pattern Recognit., 2023, pp. 9622–9631.

\bibitem{c35} Zhong Z, Rempe D, Xu D, et al. Guided conditional diffusion for controllable traffic simulation[C]//2023 IEEE International Conference on Robotics and Automation (ICRA). IEEE, 2023: 3560-3566.

\bibitem{c36} Scanlon, J. M., Kusano, K. D., Daniel, T., Alderson, C., Ogle, A., and Victor, T. Waymo simulated driving behavior in reconstructed fatal crashes within an autonomous vehicle operating domain. Accident Analysis \& Prevention, 163:106454, 2021.

\bibitem{c37} Data-driven test scenario generation for cooperative maneuver planning on highways. Applied Sciences, 10(22):8154, 2020.

\bibitem{c38} Ding, W., Chen, B., Li, B., Eun, K. J., and Zhao, D. Multimodal safety-critical scenarios generation for decision making algorithms evaluation. IEEE Robotics and Automation Letters, 6(2):1551–1558, 2021a.

\bibitem{c39} Zhang, Q., Hu, S., Sun, J., Chen, Q. A., and Mao, Z. M.
On adversarial robustness of trajectory prediction for autonomous vehicles. arXiv preprint arXiv:2201.05057, 2022.

\bibitem{c40} Arief, M., Huang, Z., Kumar, G. K. S., Bai, Y., He, S. Ding, W., Lam, H., and Zhao, D. Deep probabilistic accelerated evaluation: A certifiable rare-event simulation methodology for black-box autonomy. arXiv preprint arXiv:2006.15722, 2020.

\bibitem{c41} Anthropic, The Claude 3 Model Family: Opus, Sonnet, Haiku, https://api.semanticscholar.org/CorpusID:268232499

\bibitem{c42} Ngiam J, Caine B, Vasudevan V, et al. Scene transformer: A unified architecture for predicting multiple agent trajectories[J]. arXiv preprint arXiv:2106.08417, 2021.

\bibitem{c43} Bergamini L, Ye Y, Scheel O, et al. Simnet: Learning reactive self-driving simulations from real-world observations[C]//2021 IEEE International Conference on Robotics and Automation (ICRA). IEEE, 2021: 5119-5125.

\bibitem{c44} Wang J, Pun A, Tu J, et al. Advsim: Generating safety-critical scenarios for self-driving vehicles[C]//Proceedings of the IEEE/CVF Conference on Computer Vision and Pattern Recognition. 2021: 9909-9918.

\bibitem{c45} Xu C, Zhao D, Sangiovanni-Vincentelli A, et al. Diffscene: Diffusion-based safety-critical scenario generation for autonomous vehicles[C]//The Second Workshop on New Frontiers in Adversarial Machine Learning. 2023.

\bibitem{c46} Wilson B, Qi W, Agarwal T, et al. Argoverse 2: Next generation datasets for self-driving perception and forecasting[J]. arXiv preprint arXiv:2301.00493, 2023.

\bibitem{c47} Zhou M, Luo J, Villella J, et al. Smarts: An open-source scalable multi-agent rl training school for autonomous driving[C]//Conference on robot learning. PMLR, 2021: 264-285.

\bibitem{c48} Zhang M, Cai Z, Pan L, et al. Motiondiffuse: Text-driven human motion generation with diffusion model[J]. arXiv preprint arXiv:2208.15001, 2022.

\bibitem{c49} Suo S, Regalado S, Casas S, et al. Trafficsim: Learning to simulate realistic multi-agent behaviors[C]//Proceedings of the IEEE/CVF Conference on Computer Vision and Pattern Recognition. 2021: 10400-10409.

\bibitem{c50} Zhang K, Wang S, Jia N, et al. Integrating visual large language model and reasoning chain for driver behavior analysis and risk assessment[J]. Accident Analysis \& Prevention, 2024, 198: 107497.

\bibitem{c51} Li X, Bai Y, Cai P, et al. Towards knowledge-driven autonomous driving[J]. arXiv preprint arXiv:2312.04316, 2023.

\bibitem{c52} Cui C, Ma Y, Cao X, et al. Receive, reason, and react: Drive as you say, with large language models in autonomous vehicles[J]. IEEE Intelligent Transportation Systems Magazine, 2024.


\bibitem{c53} Tan, Shuhan, et al. "Language conditioned traffic generation." arXiv preprint arXiv:2307.07947 (2023).

\bibitem{c54} Ding, Wenhao, et al. "Realgen: Retrieval augmented generation for controllable traffic scenarios." European Conference on Computer Vision. Cham: Springer Nature Switzerland, 2024.

\bibitem{c55} Achiam, Josh, et al. "Gpt-4 technical report." arXiv preprint arXiv:2303.08774 (2023).

\bibitem{c56} Zhang, Jiawei, Chejian Xu, and Bo Li. "Chatscene: Knowledge-enabled safety-critical scenario generation for autonomous vehicles." Proceedings of the IEEE/CVF Conference on Computer Vision and Pattern Recognition. 2024.

\bibitem{c57} Cao, Yulong, et al. "Reinforcement learning with human feedback for realistic traffic simulation." 2024 IEEE International Conference on Robotics and Automation (ICRA). IEEE, 2024.

\end{thebibliography}
\end{document}